\documentclass[letterpaper, 10 pt, conference]{ieeeconf}  % Comment this line out if you need a4paper

\IEEEoverridecommandlockouts                              % This command is only needed if 

\usepackage{multicol}
\usepackage[bookmarks=false]{hyperref}
\usepackage[pdftex]{graphicx}
\usepackage{amsmath}
\usepackage{amssymb}
\usepackage{amsthm}
\usepackage{bm}
\usepackage{mathrsfs}
\usepackage{float}
\usepackage{stfloats}
\usepackage{array}
\usepackage{tabu}
\usepackage{boldline}
\usepackage{multirow}
\usepackage{hhline}
\usepackage{color, colortbl, xcolor}
\usepackage[flushleft]{threeparttable}

\newcolumntype{?}{!{\vrule width 1.5pt}}

\title{\LARGE \bf
Cascaded Gaussian Processes for \\Data-efficient Robot Dynamics Learning
}

\author{Sahand Rezaei-Shoshtari$^{1}$, David Meger$^{2}$, Inna Sharf$^{1}$
\thanks{$^{1}$Department of Mechanical Engineering, McGill University, Montreal, Canada $^{2}$Department of Computer Science, McGill University, Montreal, Canada. Correspondence to: Sahand Rezaei-Shoshtari 
        {\tt\small sahand.rezaei-shoshtari@mail.mcgill.ca}}%
}

\begin{document}

\maketitle
\thispagestyle{empty}
\pagestyle{empty}

%%%%%%%%%%%%%%%%%%%%%%%%%%%%%%%%%%%%%%%%%%%%%%%%%%%%%%%%%%%%%%
\begin{abstract}
Motivated by the recursive Newton-Euler formulation, we propose a novel cascaded Gaussian process learning framework for the inverse dynamics of robot manipulators. This approach leads to a significant dimensionality reduction which in turn results in better learning and data efficiency. We explore two formulations for the cascading: the inward and outward, both along the manipulator chain topology. The learned modeling is tested in conjunction with the classical inverse dynamics model (semi-parametric) and on its own (non-parametric) in the context of feed-forward control of the arm. Experimental results are obtained with Jaco 2 six-DOF and SARCOS seven-DOF manipulators for randomly defined sinusoidal motions of the joints in order to evaluate the performance of cascading against the standard GP learning. In addition, experiments are conducted using Jaco 2 on a task emulating a pouring maneuver. Results indicate a consistent improvement in learning speed with the inward cascaded GP model and an overall improvement in data efficiency and generalization.
\end{abstract}
%%%%%%%%%%%%%%%%%%%%%%%%%%%%%%%%%%%%%%%%%%%%%%%%%%%%%%%%%%%%%%

\section{Introduction}

Accurate inverse dynamics enables accurate control and thus several of robotics researchers have considered this problem. When the robot's rigid-body parameters are known precisely, such as through System Identification \cite{wu2010overview}, the manipulator dynamics equations can be formed, allowing us to express the inverse dynamics as a linear algebraic problem \cite{featherstone2016dynamics}. Recursive Newton-Euler (NE) algorithms have been shown to accurately and efficiently solve the inverse dynamics problem in this case. However, robots today are only imperfectly calibrated, and tend to change dynamics frequently as they interact with the world (e.g., carrying loads, using new end effectors, interacting with humans). 

Learning approaches, in principle, allow the quick adaptation of inverse dynamics to observed conditions, but present many obstacles such as the need to capture sufficient data and computational expense. Gaussian Processes (GP) have frequently been used in this context, as their non-parametric nature provides strong performance across a variety of robots with little human knowledge required \cite{nguyen2009model,nguyen2011model,vinogradska2017stability}. It is feasible to learn during live operation of robots and adapt to changing situations \cite{deisenroth2015gaussian}. The primary downside of GPs for use in robotics is their computational scaling as a function of training data size. Although many methods exist to reduce this complexity, such as sparse pseudo-inputs \cite{snelson2006sparse}, sparse spectrum GPs \cite{quia2010sparse} and Bayesian Neural Networks \cite{galMCDropout}, each of these approaches still scales with the underlying task complexity. A more intrinsically efficient representation of the inverse dynamics learning problem allows for a simpler learning and requires less data.

\begin{figure}[t!]
    \centering
    \setlength{\abovecaptionskip}{0pt}
    \includegraphics[width=0.40\textwidth]{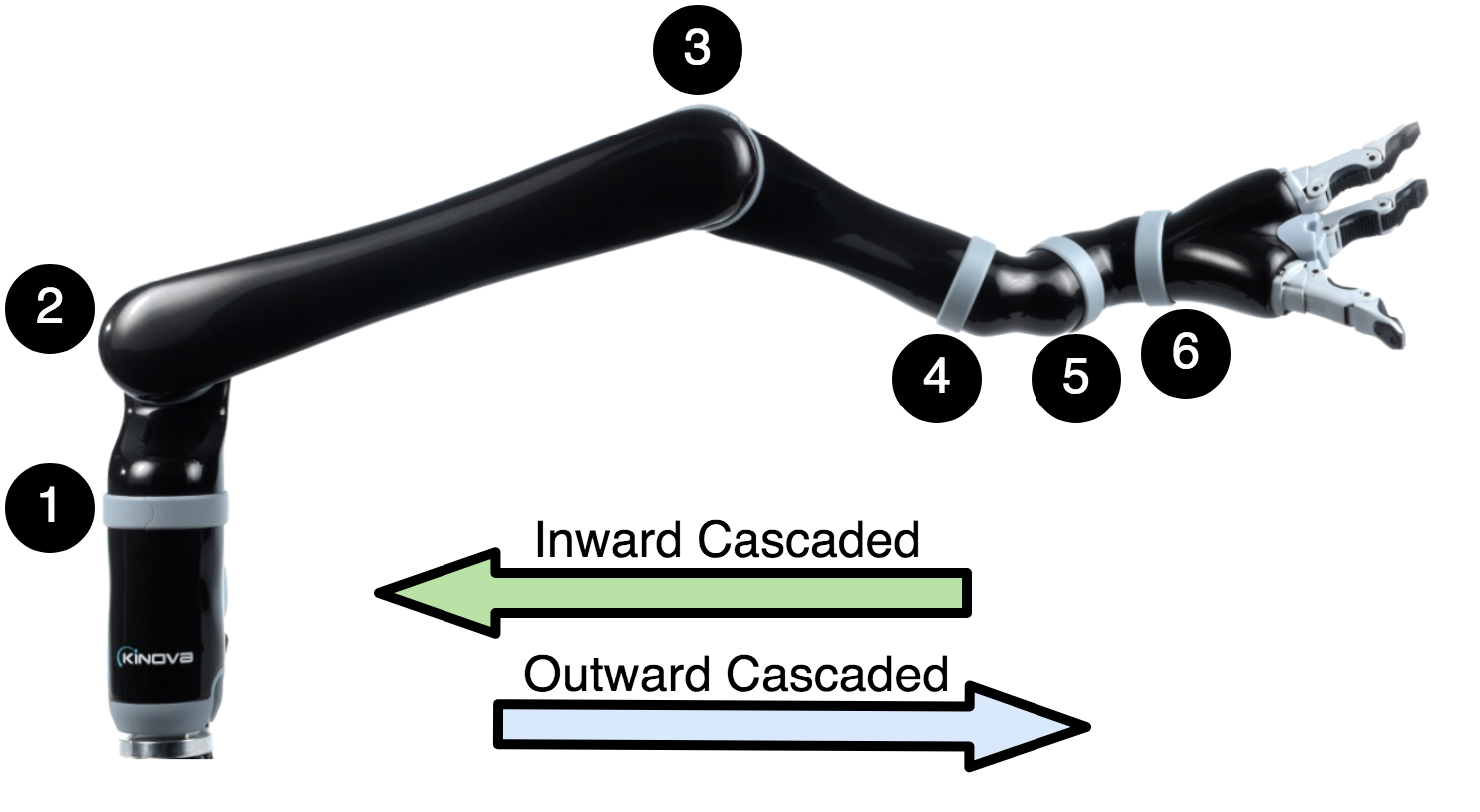}
    \caption{Implementation of outward and inward cascaded GPs on Kinova Jaco 2 robot.}
    \label{fig:overview_cascaded}
\end{figure}

This paper adapts GPs to learn the inverse dynamics of a robotic arm from data in a fashion that respects our knowledge of the underlying topology by \emph{cascading} the learning along the kinematic chain with the same factorization used in a NE solver, as illustrated in Figure \ref{fig:overview_cascaded}. Specifically, the torque output by the GP at each joint is used as an input for the subsequent joint, capturing the dynamic coupling from all previous joints. This cascaded approach utilizes many fewer input dimensions, reducing computation, but more importantly, it re-uses the results and requires learning simpler functions at each joint. Following the forward and backwards recursion of NE, we attempt both inwards and outwards cascaded versions of our GP learner.

Rather than discarding the classical inverse dynamics solution, we follow many previous approaches and combine parametric models with non-parametric models, resulting in a semi-parametric modeling framework \cite{nguyen2010using, um2014independent, romeres2016online, reinhart2017hybrid, dallasemi, wu2012semi}. Specifically, we utilize the classical inverse dynamics solution as our learner's mean function. This approach tends to achieve better generalization, data-efficiency, and faster learning due to its utilization of prior knowledge. Finally, we employ the recently-proposed derivative-free features \cite{romeres2019derivative} as alternative to the standard derivative-based variables (velocity and acceleration) as inputs to the inverse dynamics model. We evaluate our proposed approach with extensive experiments carried out on a Kinova Jaco 2 six-DOF arm and using the public SARCOS dataset \cite{vijayakumar2002statistical}.

We continue with Section II, which provides a background on model-based torque controllers and inverse dynamics learning. Section III describes the cascaded Gaussian process formulation while Section IV evaluates this method by presenting the details behind the experiments and the obtained results. Finally, section V summarizes our work.

\section{Background}
\subsection{Model-based Torque Control}
Model-free control schemes (e.g. independent-joint control and PID control) are only capable of controlling a robotic arm through set-point regulation. When used for trajectory tracking applications, these methods require intermediate set points, resulting in considerable delays and jerky motions. On the other hand, model-based controllers can make use of the inverse dynamics model to cancel out non-linearities and to achieve zero-error tracking performance \cite{chung2016motion}. 

The dynamics model of a robotic manipulator with $N$ DOF is described by Lagrange's equation of motion \cite{featherstone2016dynamics}:
\begin{equation}
    \mathbf{H}(\bm{q}) \ddot{\bm{q}} + \mathbf{C}(\bm{q}, \bm{\dot{q}}) \bm{\dot{q}} + \bm{\tau}_g(\bm{q}) = \bm{\tau} + \bm{\tau}_{\varepsilon},
    \label{eq:dynamics}
\end{equation}
where $\bm{q} \in \mathbb{R}^{N}$ is the column vector of joint positions, $\mathbf{H}(\bm q) \in \mathbb{R}^{N \times N}$ is the inertia matrix, $\mathbf{C}(\bm q, \bm{\dot{q}})\bm{\dot{q}} \in \mathbb{R}^{N}$ represents Coriolis and centrifugal forces, $\bm{\tau}_g(\bm q) \in \mathbb{R}^{N}$ denotes gravity forces, $\bm \tau \in \mathbb{R}^{N}$ is the column vector of joint control inputs to be computed, and $\bm{\tau}_\varepsilon$ is the additional torque components due to friction and disturbances. 

Feedforward torque control aims to cancel the nonlinear terms and decouple the dynamics of each link based on the theory of feedback linearization. The control signal, Equation (\ref{eq:feedforward_torque}), consists of a feedforward term computed using the inverse dynamics model and a feedback term typically on joint positions and velocities \cite{khosla1988experimental}:
\begin{align}
    \bm{\tau} &= \underbrace{\mathbf{H}(\bm{q}_\text{d}) \ddot{\bm{q}}_\text{d} + \mathbf{C}(\bm{q}_\text{d}, \bm{\dot{q}}_\text{d}) \bm{\dot{q}}_\text{d} + \bm{\tau}_g(\bm{q}_\text{d})}_{\text{Feedforward}} \nonumber \\ 
    &\hphantom{=} + \underbrace{\mathbf{K_p} (\bm{q}_\text{d} - \bm{q}) + \mathbf{K_d} (\bm{\dot{q}}_\text{d} - \bm{\dot{q}})}_{\text{Feedback}},
    \label{eq:feedforward_torque}
\end{align}
where $\bm{q}_d$ and $\bm{\dot{q}}_d$ are the desired joint positions and velocities. 

A major drawback of this parametric algorithm is the considerable degradation of controller performance in the case of an imprecise inverse dynamics model. The limitations of parametric approaches to capture friction, nonlinearities, input disturbances and robot interactions strongly motivate the need for a data-driven learning scheme.

\subsection{Gaussian Processes for Model Learning}
\label{sec:gp_lit}
Following the notation used in \cite{rasmussen2004gaussian}, a non-parametric model represented by Gaussian process regression is written as $\bm{f}(\mathbf{x}) \sim \mathcal{GP}(m(\mathbf{x}), k(\mathbf{x}, \mathbf{x'}))$, where $\mathbf{x}$, $m(\mathbf{x})$ and $k(\mathbf{x}, \mathbf{x'})$ are respectively the input, mean and covariance.

The kernel is responsible for the general shape of the sampled functions from the GP distribution and it has a large impact on the performance of the model learning. The Matern kernel is often recommended in robotics applications for two reasons. First, unlike the Squared Exponential kernel, it does not suffer from concentration of measure for high dimensional inputs and second, it is not infinitely differentiable \cite{rasmussen2004gaussian} which makes it a more realistic approach in robotics applications. The hyperparameters of the GP, lengthscale ($l$), variance ($\sigma$) and the noise variance $(\sigma_n^2)$ are obtained by maximizing the log marginal likelihood using gradient based optimization algorithms. 

While a non-parametric inverse dynamics model is a direct map from joint states ($\bm{q}, \bm{\dot{q}}, \bm{\ddot{q}}$) to joint torques, a semi-parametric model can be thought of as a combination of a rigid body dynamics (RBD) and an error term:
\begin{equation}
    \bm{\tau}(\bm{q}, \bm{\dot{q}}, \bm{\ddot{q}}) = \underbrace{\bm{\tau}_\text{RBD}(\bm{q}, \bm{\dot{q}}, \bm{\ddot{q}})}_{\text{Parametric}} + \underbrace{\bm{\varepsilon}(\bm{q}, \bm{\dot{q}}, \bm{\ddot{q}})}_{\text{Non-parametric}}.
    \label{eq:semi-parametric}
\end{equation}
Here, the parametric component, left side of Equation (\ref{eq:dynamics}), provides a baseline for the predicted torque while the non-parametric term is responsible for unmodeled dynamics of the system and the interactions of the robot. 

Assuming the robot has $N$ degrees of freedom and $\mathcal{GP}_i$ predicts the torque for joint $i$, the standard non-parametric model developed by previous authors uses $\bm{x} = [q_1, ..., q_N, \dot{q}_1, ..., \dot{q}_N, \ddot{q}_1, .., \ddot{q}_N]^T \in \mathbb{R}^{3N}$ as the input and $y = [\tau_1, ..., \tau_N] \in \mathbb{R}$ as the output. On the other hand, the non-parametric part of a  semi-parametric model keeps the same input but employs $y = [\varepsilon_1, ..., \varepsilon_N] \in \mathbb{R}$ as the output.

Proposed in \cite{nguyen2010using}, a straightforward method to integrate the RBD function into the GP framework is by setting it as the mean function, thus making the GP biased towards the prior knowledge available from the RBD model. It is often recommended to use a zero mean function for the GP, since addition of a non-zero constant to the GP posterior would negatively affect the refined uncertainty predictions \cite{rasmussen2004gaussian}. Nevertheless, we believe that the semi-parametric modeling framework has other advantages in the context of robot learning that justifies the use of non-zero mean functions. Using this approach, the torque of joint $i$ for query point $\mathbf{x_*}$ can be predicted by:
\begin{align}
    \label{eq:torque_prediction}
    \bar{\bm{\tau}}_i(\mathbf{x_*}) &= m(\mathbf{x_*}) \\
    &\hphantom{+} + k(\mathbf{x_*}, \mathbf{X}) (k(\mathbf{X}, \mathbf{X}) + \sigma_n^2 \mathbf{I})^{-1} (\mathbf{y}_i - m(\mathbf{x_*})). \nonumber
\end{align}
Equation (\ref{eq:torque_prediction}) indicates the fact that the GP is mainly responsible for modeling the error between the RBD model and actual torque measurements, because if this error $(\mathbf{y}_k - m(\mathbf{x_*}))$ is close to zero, the second term would become negligible and the predicted torque is based on the RBD function. Furthermore, according to Equation (\ref{eq:torque_prediction}), if the query point $\mathbf{x_*}$ is far from the training data $\mathbf{X}$, the resulting covariance matrix will become almost zero and the torque prediction will be a function of the RBD model only. This is in fact the reason for better generalization of semi-parametric models compared to non-parametric ones \cite{nguyen2010using}.

\section{Cascaded Gaussian Processes for Inverse Dynamics Learning}
\subsection{Cascaded Formulation}
We propose cascaded Gaussian processes as a novel data-efficient approach for inverse dynamics learning. The basic idea is to reduce the dimensions of the regression problem recursively by using the output of a GP trained on one joint as a substitute for all previous joint states, thus reducing the dimensions of the feature space. The formulation of our cascaded GP is inspired by the recursive NE algorithm for computing the inverse dynamics of robotic manipulators. The recursive NE method is comprised of two main steps; first, an outward recursion from the base of the robot to compute link velocities and accelerations based on robot kinematics and joint states. Total forces on each link are also calculated in this step using the NE equations of motion. Second, during an inward recursion, values for joint torques are computed by balancing the forces acting on each link \cite{featherstone2016dynamics}.

\begin{figure}[h]
    \centering
    \setlength{\abovecaptionskip}{-20pt}
    \includegraphics[width=0.4\textwidth]{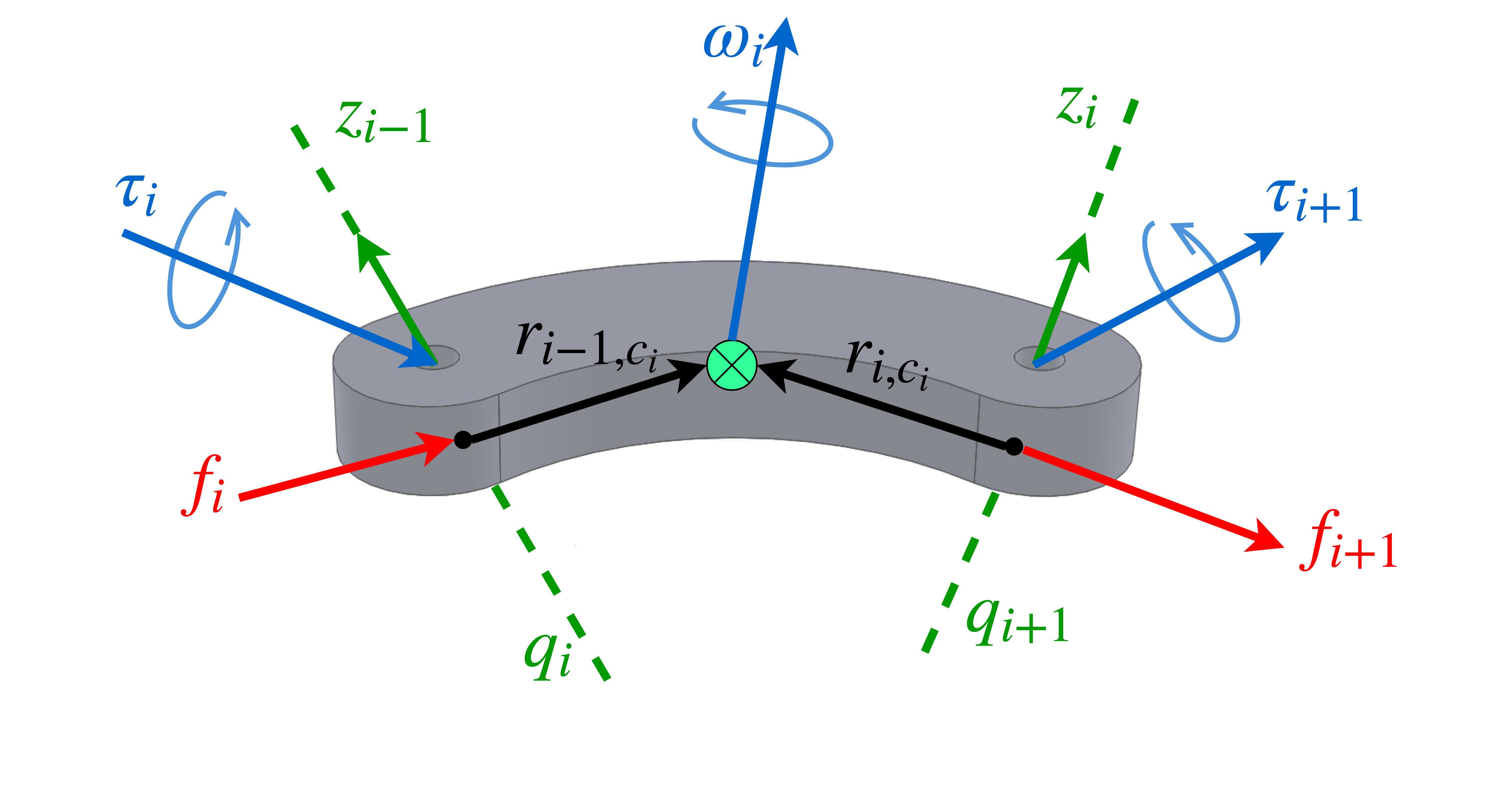}
    \caption{Free body diagram of link $i$.}
    \label{fig:recursive_NE}
\end{figure}

\begin{figure}[b!]
    \centering
    \includegraphics[width=0.48\textwidth]{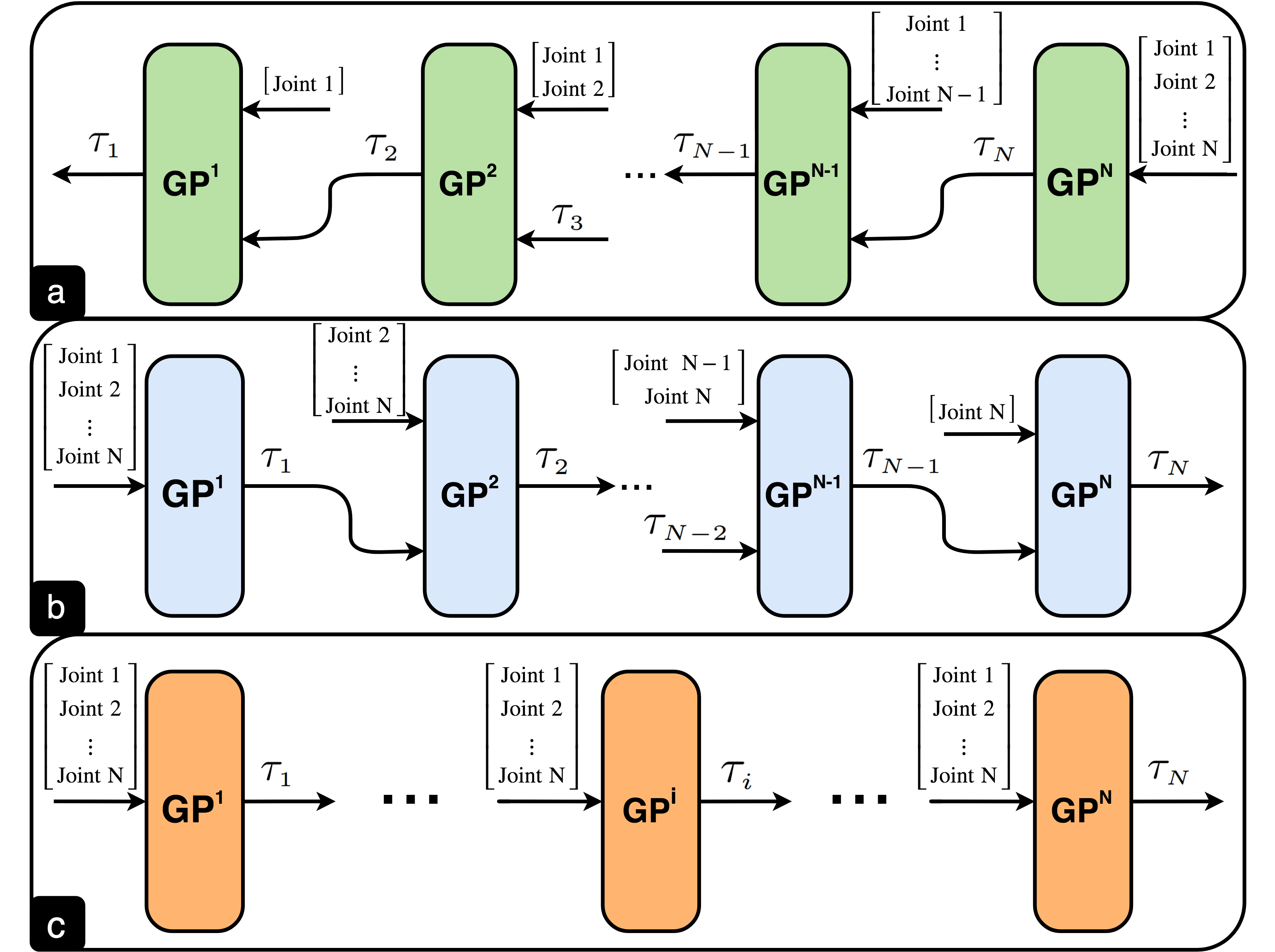}
    \caption{Overview of the proposed cascaded GP methods in comparison with the standard GP for inverse dynamics learning of robotic manipulators. \textbf{a.} Inward cascaded GP: starts from the last joint and follows an inward recursion towards the base of the robot while reducing the features. \textbf{b.} Outward cascaded GP: starts from the first joint and follows an outward recursion towards the end-effector while reducing the features. \textbf{c.} Standard GP model: trains each joint independently from the others.}
    \label{fig:models}
\end{figure}

Figure \ref{fig:recursive_NE} shows the free body diagram of link $i$; $\bm{\omega}_i$ is the angular velocity of link $i$, $z_{i}$ is the axis of joint $i$, and $\bm{f}_{i+1}$ and $\bm{\tau}_{i+1}$ are respectively the force and moment applied from link $i+1$ to link $i$. As $\bm{\omega}_i$ is calculated during the outward recursion and $f_{i+1}$ and $\bm{\tau}_{i+1}$ are available from the inward recursion step, one can solve for $\bm{f}_i$ and $\bm{\tau}_{i}$ using the Euler equations of motion:
\begin{equation}
    \begin{cases}
        \bm{f}_i &= \bm{f}_{i+1} + m_i (\bm{a}_i - \bm{g})\\
        \bm{\tau}_i &= \bm{\tau}_{i+1} - \bm{f}_{i} \times (\bm{r}_{i-1, c_i} + \bm{r}_{i, c_i}) + \bm{f}_{i+1} \times \bm{r}_{i, c_i} \\ 
        &\hphantom{=} + \mathbf{I}_i \bm{\dot{\omega}}_i + \bm{\omega}_i \times (\mathbf{I}_i \bm{\omega}_i).
    \end{cases}
    \label{eq:recursive_NE}
\end{equation}
Standard GP models (Figure \ref{fig:models}c), discussed in Section \ref{sec:gp_lit}, are trained with the states of all joints, thus resulting in a high dimensional feature space. In contrast, the proposed cascaded approach uses joint torques as a proxy for joint states. The cascading GP formulation can be implemented in either non-parametric or semi-parametric modeling paradigm.

We investigate two embodiments of the cascaded GP formulation, referred to as inward and outward cascaded GP. Our inward cascaded GP (Figure \ref{fig:models}a) is based on Equation (\ref{eq:recursive_NE}) in which during an inward recursion, the torque of joint $i+1$ $(\tau_{i+1})$ predicted by its corresponding GP can be utilized instead of the states of the outward joints $i+1$ to $N$, allowing for a reduced feature space $\bm{x} = [\tau_{i+1}, q_1, ..., q_i, \dot{q}_1, ..., \dot{q}_i, \ddot{q}_1, .., \ddot{q}_i]^T \in \mathbb{R}^{(3i+1)}$. 

The outward cascaded GP (Figure \ref{fig:models}b), on the other hand, starts from the base of the robot and reduces the dimensions of the feature space during an outward recursion procedure. Thus, the reduced feature space for joint $i$ becomes $\bm{x} = [\tau_{i-1}, q_i, ..., q_N, \dot{q}_i, ..., \dot{q}_N, \ddot{q}_i, .., \ddot{q}_N]^T \in \mathbb{R}^{3(N-i)+1}$.

The cascaded GP formulation has two main advantages over the standard inverse dynamics learning schemes. First, our method is able to generalize more effectively. According to Equation (\ref{eq:recursive_NE}), the force $\bm{f}_{i+1}$ and torque $\bm{\tau}_{i+1}$ can summarize the impact of all the outward joints; however, measuring these values requires installing force/torque sensors on each joint of a robot. Therefore, we hypothesize that a single joint torque value, which is readily available on most of the robots, can summarize the impact of all the outward joints to some extent; giving our model access to the estimate of this torque avoids the need to re-discover this pattern from training data. Compared to the regression in joint states used by standard methods, which requires observing a complete set of states covering most of the workspace, the cascaded approach better matches the topology of the arm; hence, the learning is more data-efficient and the learned model can generalize better.

Second, as a result of the reduced dimensionality of the GP inputs in the cascaded approach, several computational tasks are more efficient. During prediction, the computation of the GP scales with the number of state dimensions so our cascaded approach requires less computation. Further, optimization of the hyperparamaters becomes a significantly simpler task in lower dimensions and when the input states are more uniformly correlated to the outputs. 
% I could not fit this particular piece well
%In the context of inverse dynamics learning, Automatic Relevance Determination (ARD) is often recommended to 
%learn an individual lengthscale hyperparameter $l_d$ for each feature in order to enable the GP to determine the uncorrelated states to the output. However, the addition of hyperparameters complicates the maximization of log marginal likelihood, thus increasing the total number of iterations. Furthermore, given 
Specifically, computing the GP's marginal likelihood and its derivative scales as $(\mathcal{O}(n^3))$, with the number of training points, so the data efficiency gained by our method translates into faster learning.
%training an inverse dynamics model would take longer and become unsuitable for online implementation. On the other hand, the reduced dimension of the system introduced by cascaded approach simplifies the optimization step by reducing the number of state variables; hence, the solution converges in fewer iterations. 

\begin{figure*}[b!]
    \centering
    \includegraphics[width=\textwidth]{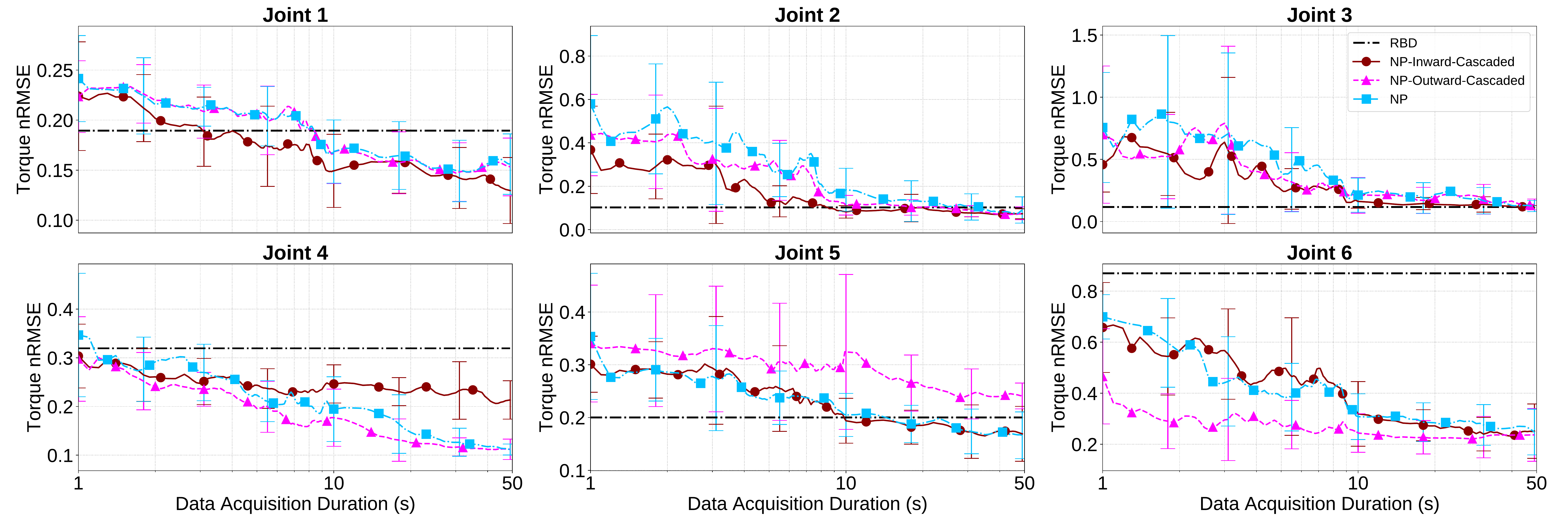}
    \caption{Normalized RMSE of torque prediction on the Jaco dataset for each joint as a function of data acquisition duration (logarithmic scale) for derivative-based non-parametric models. Training data is subsampled at 10 Hz. The evaluated models include non-parametric inward cascaded (NP-Inward-Cascaded), non-parametric outward cascaded (NP-Outward-Cascaded), non-parametric (NP) and the parametric (RBD) model. The solid lines present the nRMSE averaged over 10 subsets of training data and the error bars present the standard deviation.}
    \label{fig:prediction_np}
\end{figure*}

\subsection{Derivative-Free Features}
Utilizing derivative-free features in the context of inverse dynamics has been proposed in \cite{romeres2019derivative} as a means to address the noisy nature of numerical differentiation, an inevitable step to obtain joint velocities and accelerations. Assuming we have access to the previous $M$ joint positions, there are numerous ways to incorporate the state history as a feature. In a study reported in \cite{romeres2019derivative}, the authors concluded that derivative-free features with reduced rank provide well-rounded performance. In this method, a smaller number of features $k$ is chosen to compress the information within the position history. Physics suggests that 3 elements (position, velocity and acceleration) suffice to define a state; in addition, according to the empirical results in \cite{romeres2019derivative}, $k=3$ is the optimal choice for inverse dynamics learning. Therefore, we have set $k=3$ in this paper for the implementation of derivative-free features:
\begin{equation}
    \bm{\xi}_i = \mathbf{R} \bm{q}_i(t^\mathbf{-}),
\end{equation}
where $\bm{\xi}_i \in \mathbb{R}^{k}$ is the reduced rank derivative-free feature of joint $i$, $\bm{q}_i(t^\mathbf{-}) \in \mathbb{R}^{(M+1)}$ is the history of joint $i$ positions, and $\mathbf{R} \in \mathbb{R}^{k \times (M+1)}$ is a fully parameterised matrix. Guidelines for designing $\mathbf{R}$ can be found in \cite{romeres2019derivative}, but in the present implementation, we set $\mathbf{R} = \mathbf{I}$ and $M=2$.

\begin{figure*}[t!]
    \centering
    \includegraphics[width=\textwidth]{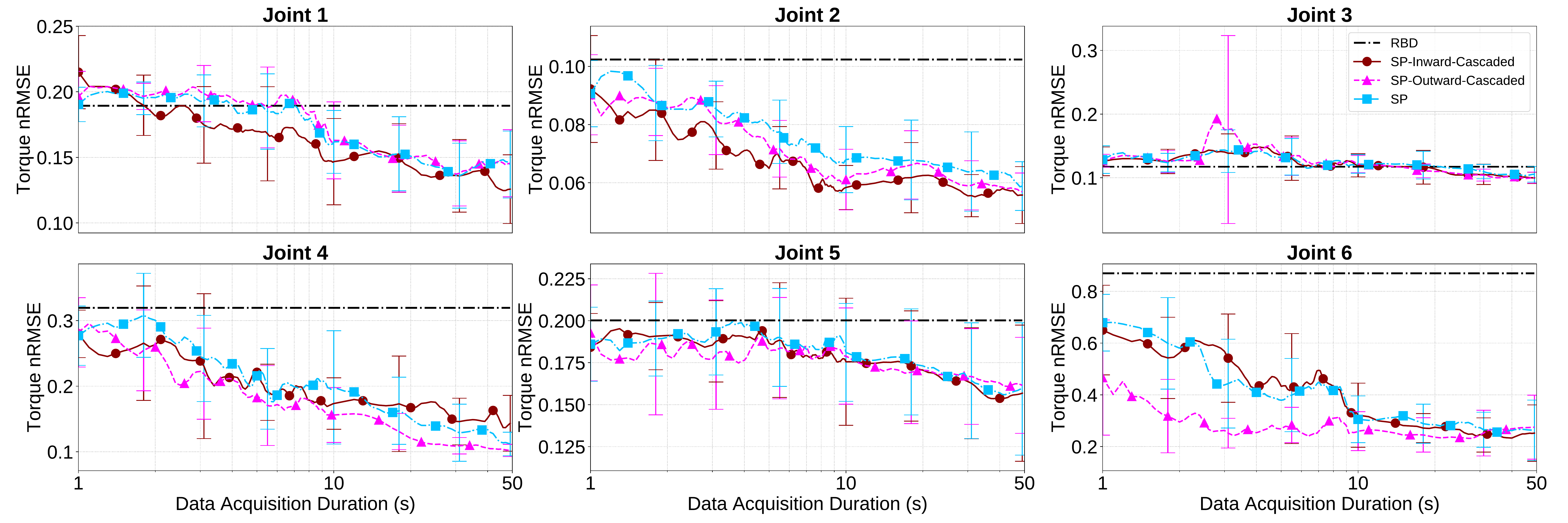}
    \caption{Normalized RMSE of torque prediction on the Jaco dataset for each joint as a function of data acquisition duration (logarithmic scale) for derivative-based semi-parametric models. Training data is subsampled at 10 Hz. The evaluated models include semi-parametric inward cascaded (SP-Inward-Cascaded), semi-parametric outward cascaded (SP-Outward-Cascaded), semi-parametric (SP) and the parametric (RBD) model. The solid lines present the nRMSE averaged over 10 subsets of training data and the error bars present the standard deviation.}
    \label{fig:prediction_sp}
\end{figure*}
\begin{table*}[b!]
    \begin{threeparttable}
        \centering
        \caption{Normalized rmse of torque prediction for derivative-based and derivative-free models on the Jaco dataset.}
        \scriptsize
\begin{tabular}[c]{V{3} m{0.1em} V{3} m{11.7em} V{3} *{6}{m{1.2em} m{1.2em} m{1.2em} V{3}}}
    \clineB{2-20}{3}
     \multicolumn{1}{c?}{\multirow{2}{*}{}} & \multicolumn{1}{c?}{\multirow{2}{*}{\textbf{Model}}} & \multicolumn{3}{c?}{\textbf{Joint 1}} & \multicolumn{3}{c?}{\textbf{Joint 2}} & \multicolumn{3}{c?}{\textbf{Joint 3}} & \multicolumn{3}{c?}{\textbf{Joint 4}} & \multicolumn{3}{c?}{\textbf{Joint 5}} & \multicolumn{3}{c?}{\textbf{Joint 6}} \\
    \cline{3-20}
    \multicolumn{1}{c?}{\multirow{2}{*}{}} & & 2s & 5s & 10s & 2s & 5s & 10s & 2s & 5s & 10s & 2s & 5s & 10s & 2s & 5s & 10s & 2s & 5s & 10s\\
    \hlineB{3}
    
    %----------------------------Derivative Based---------------------
    \multicolumn{1}{?c?}{\multirow{7}{*}{\rotatebox[origin=c]{90}{\textbf{Derivative-based}}}} & \textbf{NP-Inward-Cascaded} & \cellcolor{yellow}.219 & \cellcolor{yellow}.182 & \cellcolor{yellow}.151 & \cellcolor{yellow}.297 & \cellcolor{yellow}.178 & \cellcolor{yellow}.087 & \cellcolor{yellow}.562 & \cellcolor{yellow}.360 & \cellcolor{yellow}.161 & \cellcolor{yellow!30}.280 & .247 & .245 & \cellcolor{yellow}.288 & .253 & \cellcolor{yellow}.204 & .594 & .470 & .333\\
    \hhline{~*{19}-}
    
    & \textbf{NP-Outward-Cascaded}  & \cellcolor{yellow!30}.229 & .206 & .172 & .421 & .293 & .125 & \cellcolor{yellow!30}.565 & \cellcolor{yellow!30}.367 & .213 & \cellcolor{yellow}.271 & \cellcolor{yellow}.210 & \cellcolor{yellow}.173 & .330 & .310 & .303 & \cellcolor{yellow}.331 & \cellcolor{yellow}.286 & \cellcolor{yellow}.255\\
    \hhline{~*{19}-}
    
    & \textbf{NP} & \cellcolor{yellow!30}.229 & .207 & .169 & .488 & .351 & .174 & .769 & .553 & .212	& .298 & .236 & .201 & \cellcolor{yellow!30}.294 & \cellcolor{yellow}.239 & \cellcolor{yellow!30}.212 & .639 & .401 & .329 \\
    \clineB{2-2}{2}
    \hhline{~*{19}-}
    
    & \textbf{SP-Inward-Cascaded} & \cellcolor{yellow!30}.198 & \cellcolor{yellow}.171 & \cellcolor{yellow}.147 & \cellcolor{yellow}.085 & \cellcolor{yellow}.068 & \cellcolor{yellow}.058 & \cellcolor{yellow}.129 & .143 & \cellcolor{yellow!30}.123 & \cellcolor{yellow}.258 & .210 & .172 & \cellcolor{yellow!30}.191 & \cellcolor{yellow!30}.191 & \cellcolor{yellow}.176 & .590 & .449 & .335\\
    \hhline{~*{19}-}
    
    & \textbf{SP-Outward-Cascaded} & \cellcolor{yellow!30}.200 & .192 & .168 & \cellcolor{yellow!30}.088 & .076 & .061 & \cellcolor{yellow!30}.130 & .146	& \cellcolor{yellow!30}.125 & \cellcolor{yellow!30}.268 & \cellcolor{yellow}.192 & \cellcolor{yellow}.158 & \cellcolor{yellow}.182 & \cellcolor{yellow}.184 & \cellcolor{yellow!30}.182 & \cellcolor{yellow}.370 & \cellcolor{yellow}.268 & \cellcolor{yellow}.258\\
    \hhline{~*{19}-}
    
    & \textbf{SP} & \cellcolor{yellow}.197 & .186 & .162 & .092 & .081 & .067 & \cellcolor{yellow!30}.130 & \cellcolor{yellow}.136 & \cellcolor{yellow}.120 & .296 & .217 & .200 & \cellcolor{yellow!30}.188 & .194 & .188 & .638 & .394 & .331 \\
    \hlineB{3}
    
    %----------------------------Derivative free---------------------
    \multicolumn{1}{?c?}{\multirow{7}{*}{\rotatebox[origin=c]{90}{\textbf{Derivative-free }}}} & \textbf{NP-Inward-Cascaded-DF} & \cellcolor{yellow}.198 & \cellcolor{yellow}.189 & \cellcolor{yellow}.148 & \cellcolor{yellow}.257 & \cellcolor{yellow}.120 & \cellcolor{yellow}.097 & \cellcolor{yellow}.605 & \cellcolor{yellow!30}.283	& \cellcolor{yellow}.153 & .281 & .237 & .226 & \cellcolor{yellow}.302 & \cellcolor{yellow!30}.238 & \cellcolor{yellow}.178 & .483 & .313 & .287 \\
    \hhline{~*{19}-}
    
    & \textbf{NP-Outward-Cascaded-DF} & .211 & .200 & .172 & .351 & .145 & .126 & \cellcolor{yellow!30}.615 & \cellcolor{yellow!30}.282 & .196 & \cellcolor{yellow}.245 & \cellcolor{yellow}.200 & \cellcolor{yellow}.150 & .320 & .287 & .259 & \cellcolor{yellow}.309 & \cellcolor{yellow}.225 & \cellcolor{yellow}.257  \\
    \hhline{~*{19}-}
    
    & \textbf{NP-DF} & .214 & .204 & .175 & .435 & .144 & .141 & .705 & \cellcolor{yellow}.276 & .189 & .276 & .215 & \cellcolor{yellow!30}.153 & \cellcolor{yellow!30}.304 & \cellcolor{yellow}.228 & .189 & .429 & .304 & .294\\
    \clineB{2-2}{2}
    \hhline{~*{19}-}
    
    & \textbf{SP-Inward-Cascaded-DF}  & \cellcolor{yellow}.186 & \cellcolor{yellow}.182 & \cellcolor{yellow}.147 & \cellcolor{yellow}.076 & \cellcolor{yellow}.064 & \cellcolor{yellow!30}.067 & \cellcolor{yellow!30}.125 & \cellcolor{yellow}.124 & \cellcolor{yellow}.122 & \cellcolor{yellow}.259 & \cellcolor{yellow}.177 & .134 & \cellcolor{yellow!30}.194 & .196 & .177 & .471 & .314 & .294\\
    \hhline{~*{19}-}
    
    & \textbf{SP-Outward-Cascaded-DF} & \cellcolor{yellow!30}.195 & \cellcolor{yellow}.182 & .161 & \cellcolor{yellow}.076 & .068 & \cellcolor{yellow}.065 & \cellcolor{yellow!30}.122  & \cellcolor{yellow!30}.129 & \cellcolor{yellow!30}.126 & \cellcolor{yellow!30}.261 & .186 & \cellcolor{yellow}.121 & \cellcolor{yellow}.186 & \cellcolor{yellow}.177 & \cellcolor{yellow}.167 & \cellcolor{yellow}.329 & \cellcolor{yellow}.232 & \cellcolor{yellow}.258 \\
    \hhline{~*{19}-}
    
    & \textbf{SP-DF} & .196 & \cellcolor{yellow!30}.186 & .165 & \cellcolor{yellow!30}.079 & .072 & \cellcolor{yellow!30}.068 & \cellcolor{yellow}.121 & \cellcolor{yellow}.124 & \cellcolor{yellow!30}.125 & .285 & .216 & .133 & \cellcolor{yellow!30}.192 & .192 & \cellcolor{yellow!30}.171 & .431 & .300 & .292\\
    
    \hlineB{3}
\end{tabular}
        \label{table:tab_results}   
        \begin{tablenotes}
            \item The values report an average of error over a 1-second interval (e.g. 1-2 second, 4-5 second, 9-10 second) of training data. The yellow highlighted regions present the lowest error among the models of the same category (non-parametric and semi-parametric), and the light yellow highlighted regions indicate an almost similar performance (up to 5 percent difference) among the models.
        \end{tablenotes}
    \end{threeparttable}
\end{table*}

\section{Evaluation}
In this section, we evaluate the performance of the proposed cascaded Gaussian process approach in terms of data-efficiency, generalization and learning efficiency. Two evaluation scenarios are considered here: first, a quantitative evaluation with a standard training-testing procedure on subsets of a dataset. Second, a qualitative evaluation using task-based training episodes in which the robot attempts to learn the dynamics of various tasks while the data is transferred from one to another.

\subsection{Setup}
\label{sec:setup}
The experiments are carried out on Kinova Jaco 2 robot, a lightweight six-DOF robotic manipulator with a maximum payload of 1.6 kg and a reach of 90 cm. The robot is equipped with joint encoders and joint torque sensors, and it can be monitored and commanded through a ROS (Robot Operating System) \cite{quigley2009ros} package. The proposed algorithms as well as the feedforward torque controller have been implemented in Python as a ROS package on top of Kinova API. Due to the connectivity limitations of the robot, the torque controller runs at 40 Hz. Prior to applying the controller on the actual robot, it has been thoroughly tested in Gazebo, an open-source simulation environment \cite{koenig2004design}.

For implementation of our cascaded Gaussian process framework, we used GPy\footnote{\href{https://github.com/SheffieldML/GPy}{https://github.com/SheffieldML/GPy}}, an open-source Gaussian process library in Python. In addition, PyBullet\footnote{\href{https://github.com/bulletphysics/bullet3}{https://github.com/bulletphysics/bullet3}}, a Python implementation of Bullet Physics Engine, has been utilized for parametric inverse dynamics modeling of Kinova Jaco 2 arm. Our model learning framework learns offline but predicts the torques online.

\subsection{General Evaluation}
\label{sec:eval_results}
The aim of this set of experiments is to provide a quantitative evaluation of data-efficiency and accuracy of our proposed cascaded method. Two datasets have been used:
\begin{itemize}
    \item Our own dataset collected on Kinova Jaco 2 arm and referred to as the Jaco dataset.
    \item The publicly available SARCOS dataset \footnote{\href{http://www.gaussianprocess.org/gpml/data/}{SARCOS dataset is available at http://www.gaussianprocess.org/gpml/data/}} \cite{vijayakumar2002statistical}.
\end{itemize}
Every model variant has been evaluated on the Jaco dataset. However, due to the discrepancies found in some of the results, SARCOS dataset was also used to evaluate the non-parametric models.

\subsubsection{Training}
In the Jaco dataset, the training set consists of 600 seconds of velocity controlled robot motion subsampled at 10 Hz and divided into 10 subsets. The motion for each subset has been generated by commanding each joint rate as a summation of multiple sinusoidal functions with random amplitudes, frequencies and phase angles. Therefore, the training set ensures a wide variety of joint states and robot configurations. Moreover, the test dataset consists of three one-minute motions, distinct from the training set, each collected with randomly sampled amplitudes, frequencies and phase angles.

The SARCOS dataset consists of 50,000 data points collected at 50 Hz from various movement patterns on the seven-DOF anthropomorphic SARCOS robot. The test set includes 10 percent of the data. For consistency with the Jaco dataset, we have subsampled the SARCOS dataset at 10 Hz and divided the training set into 9 subsets. 

\subsubsection{Results}
Since our main goal is to improve the data-efficiency of dynamics learning, we train our model on a series of ever-increasing datasets, evaluating each resulting learned model on the test set. This allows evaluation of how much data each method requires before achieving low error.

\begin{figure*}[t!]
    \centering
    \includegraphics[width=\textwidth]{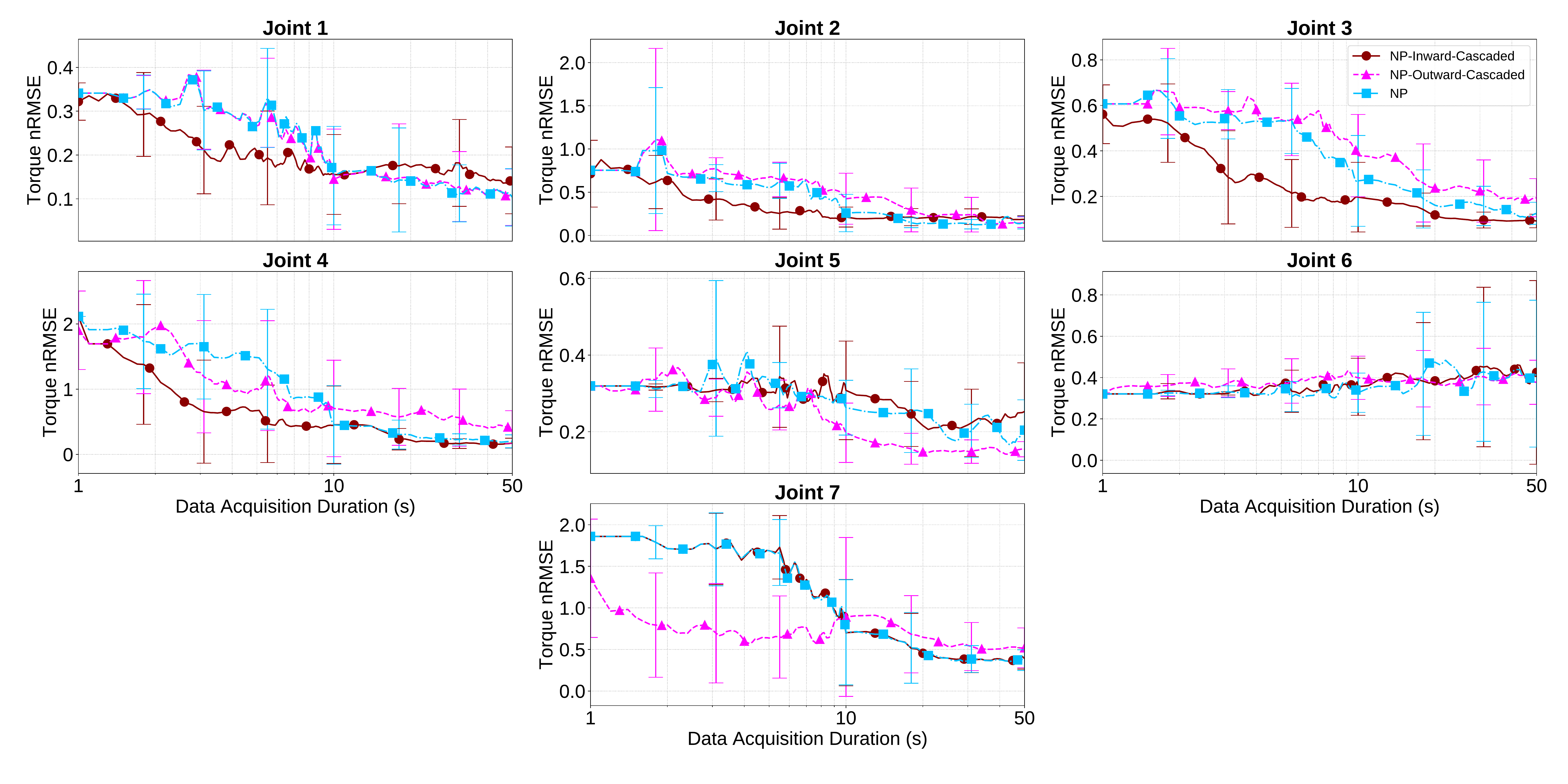}
    \caption{Normalized RMSE of torque prediction on the SARCOS dataset for each joint as a function of data acquisition duration (logarithmic scale) for derivative-based non-parametric models. Training data is subsampled at 10 Hz. The evaluated models include non-parametric inward cascaded (NP-Inward-Cascaded), non-parametric outward cascaded (NP-Outward-Cascaded) and the non-parametric (NP) model. The solid lines present the nRMSE averaged over 9 subsets of training data and the error bars present the standard deviation.}
    \label{fig:prediction_sarcos}
\end{figure*}

Table \ref{table:tab_results} shows the evaluated normalized root-mean-square error (nRMSE = RMSE/range) of torque predictions on the Jaco dataset of all models for three time stamps (averaged on ranges 1-2s, 4-5s, and 9-10s) of training time, whereas Figures \ref{fig:prediction_np} and \ref{fig:prediction_sp} present the nRMSE of torque prediction as a function of data acquisition duration for the first 50 seconds of the data acquisition for non-parametric and semi-parametric models, evaluated on the Jaco dataset. Finally, Figure \ref{fig:prediction_sarcos} reports the results obtained from non-parametric models on the SARCOS dataset.

The results show that in the case of non-parametric models (Figures \ref{fig:prediction_np} and \ref{fig:prediction_sarcos}), cascaded models in general outperform standard models in terms of data efficiency, even though these models have access to a far smaller set of robot states. Overall, the inward cascaded model, which is consistent with the direction of the recursive NE method solution for the inverse dynamics torques, provides more accurate predictions compared to the outward cascaded. 

\begin{figure}[h]
    \centering
    \setlength{\abovecaptionskip}{0pt}
    \includegraphics[width=0.37\textwidth]{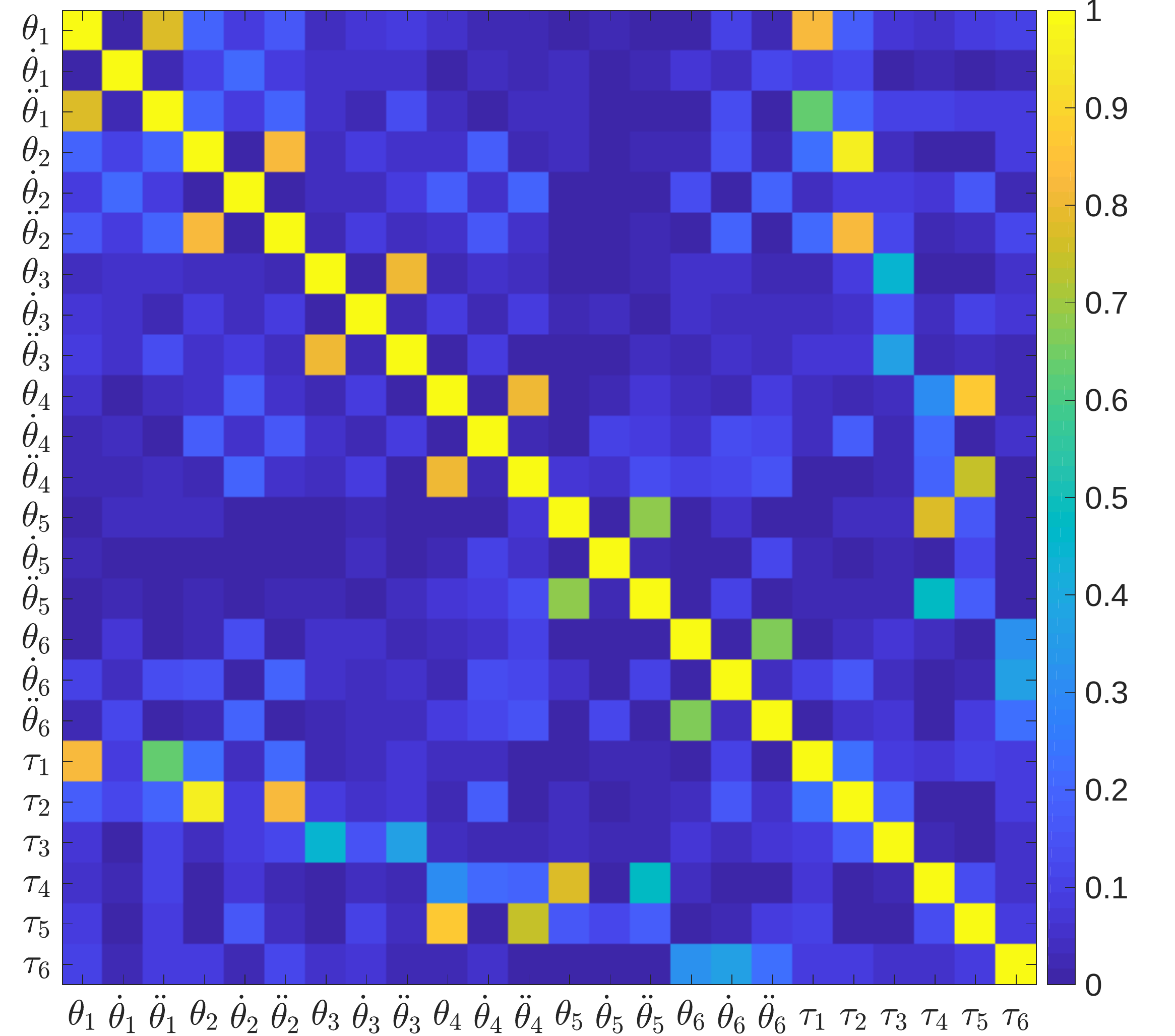}
    \caption{Correlation matrix of robot joint states over the dataset. The plot reports the absolute values of Spearman's rank correlation coefficient.}
    \label{fig:correlation}
\end{figure}

Nonetheless, outward cascaded model outperforms its inward counterpart on joint 4 in the Jaco dataset and the last joint in both of the datasets. The discrepancy of joint 4 can be explained by the intrinsic properties of the Jaco 2 arm. As Figure \ref{fig:correlation} suggests, due to the strong coupling between joints 4 and 5, the torque of each joint is more correlated to the other rather that its own states; therefore, not having access to states of joint 5, the inward cascaded model yields less accurate predictions for joint 4. The fact that this discrepancy is not present in the results obtained from the SARCOS dataset (Figure \ref{fig:prediction_sarcos}), confirms the data correlation analysis. 

The discrepancy of the last joint can be explained by the fact that the torque on the wrist of the robot is only a function of its own states, as it is the last joint of the open-chain robot arm topology. Hence, the outward cascaded model, which has access only to the states of this joint and torque of the previous one, learns the torque in a much smaller feature space, resulting in more accurate and data-efficient predictions. 

The same general trends can be observed in the case of semi-parametric models (Figure \ref{fig:prediction_sp}). However, the improvement provided by cascaded modeling is less significant as the rigid body dynamics model used as the mean function already reduces learning complexity to some extent. The comments made earlier regarding joints 4 and 6 on the Jaco dataset and joint 7 on the SARCOS dataset apply to semi-parametric models as well.

Finally, as already addressed in \cite{romeres2019derivative}, derivative-free features (Table \ref{table:tab_results}) help with the transient response of the GP models (i.e., the prediction accuracy of the model within the first seconds of training data), as they do not rely on the noisy and delayed nature of numerical differentiation. However, this improvement is limited to the first few seconds of data acquisition and as the number of data points increases, derivative-free features lack the generalization of derivative-based features. 


\begin{figure}[b!]
    \centering
    \includegraphics[width=0.5\textwidth]{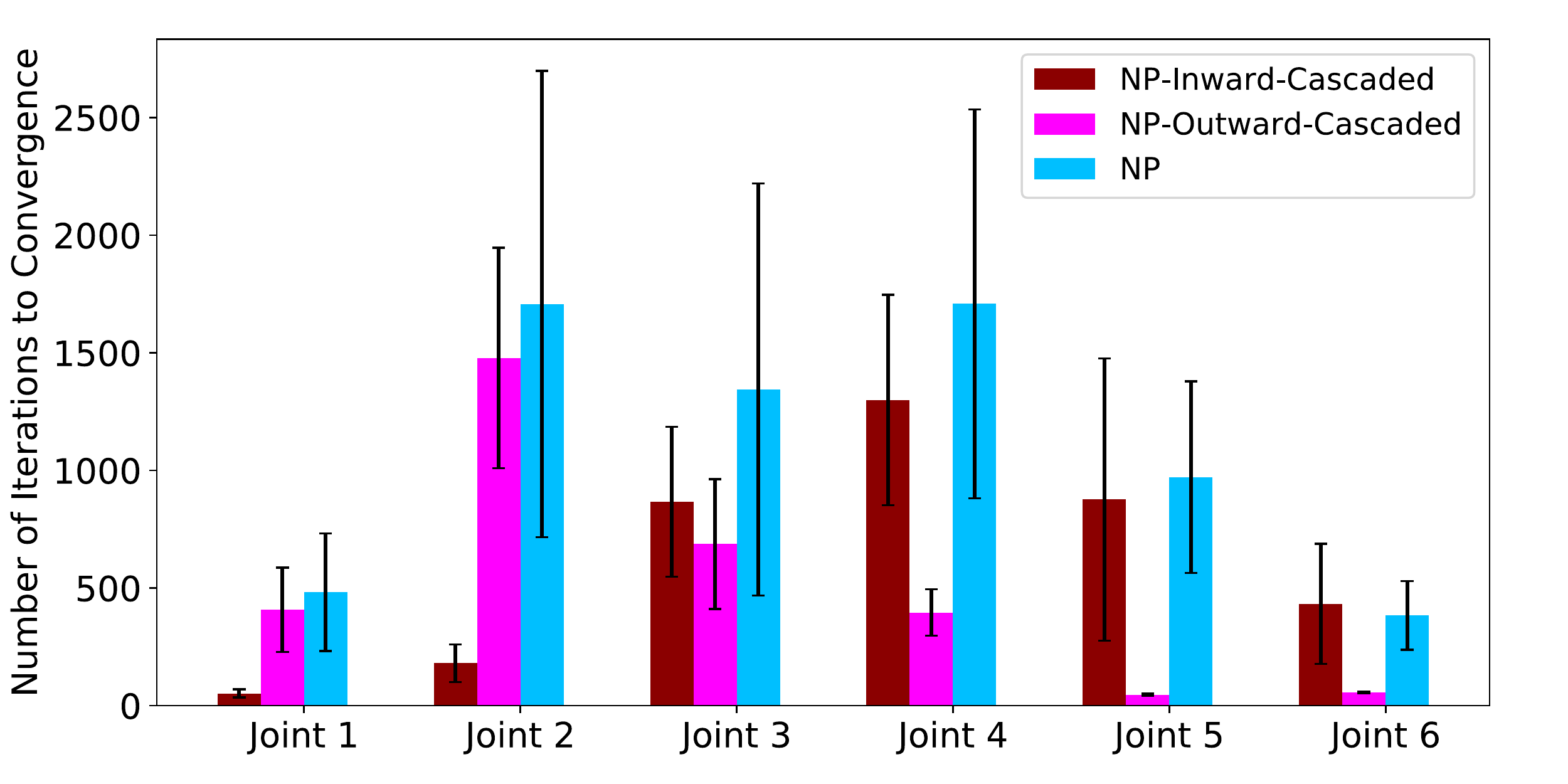}
    \caption{Comparison of number of iterations of outward and inward cascaded models with the standard model averaged over 10 restarts of the optimizer. The log marginal likelihood optimization is converged once the maximum projected gradient is less than $1\mathrm{e}\scriptstyle{-}\displaystyle{5}$ or the normalized change in the objective function is less than $2\mathrm{e}\scriptstyle{-}\displaystyle{9}$.}
    \label{fig:run_time}
\end{figure}
\subsection{Learning Efficiency}
The hyperparameters of a Gaussian process are most often obtained by maximizing the log marginal likelihood. Gradient based methods are usually used to solve this non-convex optimization problem. In high-dimensional systems the solution may converge to a local optimum that is far from the ideal solution. In order to prevent this, multiple restarts of the optimization procedure are required. Furthermore, since computing both the log marginal likelihood and its gradient has the computational complexity of $\mathcal{O}(n^3)$, solving for the optimal hyperparameters requires a significant amount of time for large datasets. 

First, the data-efficiency of the cascaded approach helps to decrease the size of the dataset required to learn a reliable model and thus decreases the training time significantly. Second, as a result of the reduced dimensionality of the cascaded GP, the total number of hyperparameters is decreased, leading to a simpler optimization problem. Consequently, log marginal likelihood is maximized with significantly fewer iterations. Figure \ref{fig:run_time} compares the average number of iterations to train cascaded and standard GP models with 500 data points. As expected, cascaded models converge quicker on the joints with smaller feature state: joints closer to the base for the inward cascaded and joints closer to the end-effector for the outward cascaded model.

\subsection{Task Based Evaluation}
As a qualitative evaluation of our proposed cascaded modeling framework, we have applied it to a series of ``pouring tasks'' in which the torque controlled robot is required to move a bottle and tip it to allow marbles to roll out of the bottle into a jar. Each task attempts to explore a different part of the workspace while the data and the learned model is transferred between the tasks. As the parametric model is not aware of the unknown load carried by the robot, the baseline performance is limited. However, cascaded GP modeling proposed here is capable of learning the dynamics of each task within the first few attempts and it can generalize well to the tasks it has already observed. The results can be seen in the supplementary video \footnote{\href{https://youtu.be/il-wWuyRZIY}{https://youtu.be/il-wWuyRZIY}}.

% **************************************** 

\section{Conclusion}
In this paper, we introduced the cascaded Gaussian process as a novel approach for inverse dynamics learning. The proposed method actively reduces the number of dimensions of the feature space by substituting joint states with joint torques. We explored two formulations of the cascaded Gaussian process, namely inward and outward models. In general, the inward models outperform the outward models as suggested by their consistency with the recursive Newton-Euler method. Overall, cascaded models provide a more data-efficient and faster training compared to the standard models; this improvement is more evident in the case of non-parametric models. Moreover, experimental results for a pouring task confirm the data-efficiency of the cascaded modeling.

As a final wrap-up of the study on the variations of inverse dynamic model learning schemes, we would suggest semi-parametric cascaded GPs with derivative-based features. The semi-parametric variant and cascaded approach both help with the generalization and data-efficiency of the model. Although derivative-free features improve the consistency of the results by omitting the inherent noise present in the differentiation procedure, this improvement is limited to the first few seconds of data acquisition and comes with the cost of reducing the generalization of the learned model.

\section*{Acknowledgments}
This work was supported by the National Sciences and Engineering Research Council (NSERC) Canadian Robotics Network (NCRN).

\bibliographystyle{IEEEtran.bst}
\bibliography{IEEEabrv.bib,references}

\begin{thebibliography}{10}
\providecommand{\url}[1]{#1}
\csname url@rmstyle\endcsname
\providecommand{\newblock}{\relax}
\providecommand{\bibinfo}[2]{#2}
\providecommand\BIBentrySTDinterwordspacing{\spaceskip=0pt\relax}
\providecommand\BIBentryALTinterwordstretchfactor{4}
\providecommand\BIBentryALTinterwordspacing{\spaceskip=\fontdimen2\font plus
\BIBentryALTinterwordstretchfactor\fontdimen3\font minus
  \fontdimen4\font\relax}
\providecommand\BIBforeignlanguage[2]{{%
\expandafter\ifx\csname l@#1\endcsname\relax
\typeout{** WARNING: IEEEtran.bst: No hyphenation pattern has been}%
\typeout{** loaded for the language `#1'. Using the pattern for}%
\typeout{** the default language instead.}%
\else
\language=\csname l@#1\endcsname
\fi
#2}}

\bibitem{wu2010overview}
J.~Wu, J.~Wang, and Z.~You, ``An overview of dynamic parameter identification
  of robots,'' \emph{Robotics and computer-integrated manufacturing}, vol.~26,
  no.~5, pp. 414--419, 2010.

\bibitem{featherstone2016dynamics}
R.~Featherstone and D.~E. Orin, ``Dynamics,'' in \emph{Springer Handbook of
  Robotics}.\hskip 1em plus 0.5em minus 0.4em\relax Springer, 2016, pp. 37--66.

\bibitem{nguyen2009model}
D.~Nguyen-Tuong, M.~Seeger, and J.~Peters, ``Model learning with local gaussian
  process regression,'' \emph{Advanced Robotics}, vol.~23, no.~15, pp.
  2015--2034, 2009.

\bibitem{nguyen2011model}
D.~Nguyen-Tuong and J.~Peters, ``Model learning for robot control: a survey,''
  \emph{Cognitive processing}, vol.~12, no.~4, pp. 319--340, 2011.

\bibitem{vinogradska2017stability}
J.~Vinogradska, B.~Bischoff, D.~Nguyen-Tuong, and J.~Peters, ``Stability of
  controllers for gaussian process dynamics,'' \emph{The Journal of Machine
  Learning Research}, vol.~18, no.~1, pp. 3483--3519, 2017.

\bibitem{deisenroth2015gaussian}
M.~P. Deisenroth, D.~Fox, and C.~E. Rasmussen, ``Gaussian processes for
  data-efficient learning in robotics and control,'' \emph{IEEE transactions on
  pattern analysis and machine intelligence}, vol.~37, no.~2, pp. 408--423,
  2015.

\bibitem{snelson2006sparse}
E.~Snelson and Z.~Ghahramani, ``Sparse gaussian processes using
  pseudo-inputs,'' in \emph{Advances in neural information processing systems},
  2006, pp. 1257--1264.

\bibitem{quia2010sparse}
J.~Qui\~nonero Candela, C.~E. Rasmussen, A.~R. Figueiras-Vidal, \emph{et~al.},
  ``Sparse spectrum gaussian process regression,'' \emph{Journal of Machine
  Learning Research}, vol.~11, no. Jun, pp. 1865--1881, 2010.

\bibitem{galMCDropout}
Y.~Gal and Z.~Ghahramani, ``Dropout as a bayesian approximation: Representing
  model uncertainty in deep learning,'' in \emph{Proceedings of the 33rd
  International Conference on Machine Learning (ICML)}, 2016.

\bibitem{nguyen2010using}
D.~Nguyen-Tuong and J.~Peters, ``Using model knowledge for learning inverse
  dynamics.'' in \emph{Robotics and Automation (ICRA), 2010 IEEE International
  Conference on}.\hskip 1em plus 0.5em minus 0.4em\relax IEEE, 2010, pp.
  2677--2682.

\bibitem{um2014independent}
T.~T. Um, M.~S. Park, and J.-M. Park, ``Independent joint learning: A novel
  task-to-task transfer learning scheme for robot models,'' in \emph{Robotics
  and Automation (ICRA), 2014 IEEE International Conference on}.\hskip 1em plus
  0.5em minus 0.4em\relax IEEE, 2014, pp. 5679--5684.

\bibitem{romeres2016online}
D.~Romeres, M.~Zorzi, R.~Camoriano, and A.~Chiuso, ``Online semi-parametric
  learning for inverse dynamics modeling,'' in \emph{Decision and Control
  (CDC), 2016 IEEE 55th Conference on}.\hskip 1em plus 0.5em minus 0.4em\relax
  IEEE, 2016, pp. 2945--2950.

\bibitem{reinhart2017hybrid}
R.~F. Reinhart, Z.~Shareef, and J.~J. Steil, ``Hybrid analytical and
  data-driven modeling for feed-forward robot control,'' \emph{Sensors},
  vol.~17, no.~2, p. 311, 2017.

\bibitem{dallasemi}
A.~Dalla~Libera, E.~Tosello, R.~Carli, G.~Pillonetto, and S.~Ghidoni, ``A
  semi-parametric robot inverse dynamics estimator with friction compensation
  for human-robot interaction,'' in \emph{Intelligent Robots and Systems
  (IROS), 2018 IEEE/RSJ International Conference on}.\hskip 1em plus 0.5em
  minus 0.4em\relax IEEE, 2018.

\bibitem{wu2012semi}
T.~Wu and J.~Movellan, ``Semi-parametric gaussian process for robot system
  identification,'' in \emph{Intelligent Robots and Systems (IROS), 2012
  IEEE/RSJ International Conference on}.\hskip 1em plus 0.5em minus 0.4em\relax
  IEEE, 2012, pp. 725--731.

\bibitem{romeres2019derivative}
D.~Romeres, M.~Zorzi, R.~Camoriano, S.~Traversaro, and A.~Chiuso,
  ``Derivative-free online learning of inverse dynamics models,'' \emph{IEEE
  Transactions on Control Systems Technology}, 2019.

\bibitem{vijayakumar2002statistical}
S.~Vijayakumar, A.~D'souza, T.~Shibata, J.~Conradt, and S.~Schaal,
  ``Statistical learning for humanoid robots,'' \emph{Autonomous Robots},
  vol.~12, no.~1, pp. 55--69, 2002.

\bibitem{chung2016motion}
W.~K. Chung, L.-C. Fu, and T.~Kr{\"o}ger, ``Motion control,'' in \emph{Springer
  handbook of robotics}.\hskip 1em plus 0.5em minus 0.4em\relax Springer, 2016,
  pp. 163--194.

\bibitem{khosla1988experimental}
P.~K. Khosla and T.~Kanade, ``Experimental evaluation of nonlinear feedback and
  feedforward control schemes for manipulators,'' \emph{The International
  Journal of Robotics Research}, vol.~7, no.~1, pp. 18--28, 1988.

\bibitem{rasmussen2004gaussian}
C.~E. Rasmussen, ``Gaussian processes in machine learning,'' in \emph{Advanced
  lectures on machine learning}.\hskip 1em plus 0.5em minus 0.4em\relax
  Springer, 2004, pp. 63--71.

\bibitem{quigley2009ros}
M.~Quigley, K.~Conley, B.~Gerkey, J.~Faust, T.~Foote, J.~Leibs, R.~Wheeler, and
  A.~Y. Ng, ``Ros: an open-source robot operating system,'' in \emph{ICRA
  workshop on open source software}, vol.~3, no. 3.2.\hskip 1em plus 0.5em
  minus 0.4em\relax Kobe, Japan, 2009, p.~5.

\bibitem{koenig2004design}
N.~P. Koenig and A.~Howard, ``Design and use paradigms for gazebo, an
  open-source multi-robot simulator.'' in \emph{Intelligent Robots and Systems
  (IROS), 2004 IEEE/RSJ International Conference on}.\hskip 1em plus 0.5em
  minus 0.4em\relax IEEE, 2004, pp. 2149--2154.

\end{thebibliography}

\end{document}